# Pneumatic Modelling for Adroit Manipulation Platform

Visak CV[1*] and Vikash Kumar[2*]

*Abstract*— ADROIT Manipulation platform is a pneumatically actuated, tendon driven 28 degree of freedom platform being developed for investigating complex hand manipulation behaviors. ADROIT derives its unique capabilities, necessary to support dynamic and dexterous manipulation, from a custom designed high performance pneumatic actuation system for tendon driven hands. The custom pneumatic actuation system is fast, strong, low friction-stiction, compliant and is capable to actuating a shadow hand skeleton faster that human capabilities – at a unique combination of speed, force and compliance that has never been achieved before. In this paper, we develop models for the pneumatic muscles of ADROIT and perform a thorough investigation of the various parameters that affect pressure dynamics in a pneumatic system such as, different cylinder types, leakage from valves and cylinders, valve deadzone, input pressure fluctuations etc to improve the model's accuracy.

## I. INTRODUCTION

Adroit manipulation platform [1] is a 28 dof robotic system being developed with an aim to achieve complex human-like object manipulation. It consists of a 24 dof bio mimetic hand and a 4 dof robotic arm. The robot is pneumatically actuated and it owes its unique capabilities such as speed, strength and compliance to the custom actuation structure of the robot, each joint is antagonistically actuated through tendon transmission, similar to antagonistic muscles in the human body.

Pneumatically actuated robots are desirable yet hard to control. The compressibility of air makes the pressure dynamics highly non linear and reduces the bandwidth of the over all system. Pneumatic actuators are controlled using a pneumatic valve which controls the rate of change of pressure (by controlling the orifice area) inside the cylinder chamber, thereby making the whole system follow third order dynamics. Furthermore, the pressure dynamics is also affected by undesirable factors such as valve dead-zone, air leakage from the valves and cylinders, delays from the connecting tube lengths and input pressure variations. The difficulty in accounting for these factors have led to a limited use of pneumatic actuators in robotic applications. Despite these drawbacks, pneumatic actuators are still desirable because they are clean, have a lower specific weight and a higher power rate than an equivalent electromechanical actuator. They are easy to maintain and handle. These factors make it worthwhile to explore pneumatic actuators as a viable actuation system for robotics.

As processors are getting faster, model based trajectory optimization techniques are increasingly being deployed to handle nonlinear systems [2]. These techniques leverage the model of the system to see through the planning horizon and deliver a locally optimal policy. The strength of trajectory optimization techniques lies in the predictive capability of the model of the system. Fast update of the policy enables it to handle non-linearities and modelling discrepancies. In order to tame the non-linearity and large timescale of pneumatic system, we plan to leverage the strengths of online trajectory optimization to build an effective low level controller that hides the complications of pneumatics and abstracts out a simple force actuator to the user. The goal of this work is to perform a thorough investigation across relevant parameters to develop models for ADROIT's pneumatic actuators that is robust to fast control signals and aggressive volume changes. This work strictly focuses on developing pneumatic models our of system. In [3] we leverage the models developed here to realize a model based high-performance pneumatic controller.

In this paper, we use the thin-port pneumatic model, which is a popular method of modelling pneumatic drives derived from the principles of thermodynamics. We present the pressure modelling results on different pneumatic cylinders and valves, compare the performance of each of them to validate the method adopted. Since the actuation system of Adroit was inspired by Kokoro's DiegoSan [4] humanoid robot, we use simple 2-dof shoulder joint of the DiegoSan robot as our second testbed for model validation.

In section II we discuss some of the previous research that has been done with pneumatic actuators. Modelling of the pressure dynamics is discussed in section III. In section IV, a brief description of all the factors that affects pressure dynamics is given. In section V, an overview of the hardware used and the control architecture for Adroit is provided. In section VI, the experimental method is described. The pressure modelling results with different hardware combinations is presented as a validation of the method adopted. Finally, we discuss the performance of the model and highlight the factors that affect the accuracy of the model.

## II. RELATED WORK

Physics based mathematical model of a pneumatic system is developed from three equations, the ideal gas law, conservation of mass and the energy equation. The most popular assumption among researchers has been to neglect the temperature dynamics and consider the expansion and compression of air as a thermodynamic process with the specific heat coefficient $n$ varying from $1-1.4$, derivation of these pneumatic models is described in [5]. Researchers have successfully developed models based on assumption that

Authors are with the Department of Computer Science & Engineering, University of Washington, USA. Email:visakc@uw.edu[1], vikash@cs.washington.edu[2]. Authors contributed equally* the process is either isothermal ($n=1$), isentropic ($n=1.4$) or a polytropic process ($1 < n < 1.4$). Carneiro et al. [6] reviews these different approaches to model the thermodynamic process.

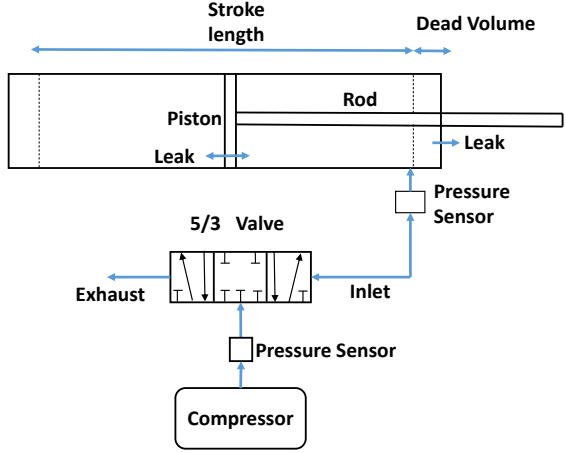

**Fig. 1:** A schematic diagram of a single actuated pneumatic cylinder controlled using a 5/3 proportional valve.

Richer et al. [7], reported slightly different approach to modelling the pressure dynamics by considering the expansion process as adiabatic and the compression as isothermal, hence having different specific heat coefficients for mass flow into and out of the chamber. In Gulati et al, [8] a force error based Lyapunov pressure observer was designed considering both the charging and discharging process as isothermal, In Gulati et al [9], the pressure observer was used alongside a sliding mode controller. In both these works, the pressure estimates was tested for simple trajectories like sinusoidal and square waves. Pandian et al. [10] also presented a pressure observer and sliding mode controller, but again the observer was only tested for simple reference pressure trajectories.

Xue et al. [11] developed a controlled auto-regressive moving average (CARMA) pneumatics model based on theoretical analysis of the pressure dynamics. Tassa et al. [2] developed a non linear parametric model for the pressure dynamics based on experimental work on a humanoid robot. In most of the previous work on modelling pneumatic actuators, researchers have refrained from commenting on the model's performance with factors such as cylinder volume, cylinder leakage and fast piston movements. In this work, we investigate the effect of these parameters and focus on developing a model that can predict pressure changes for rapid command inputs and volume changes, something that has not been reported before and is important for robotic applications.

## III. PRESSURE DYNAMICS

A schematic diagram of a pneumatic system is shown in Figure 1.

### A. Thin-Port model

In a pneumatic system, the valve essentially controls the orifice area through which compressed air flows into the chamber of the actuator. In the thin port model, area of the orifice is assumed to be small and also the plate through which the air flows from higher pressure region to lower pressure is considered to be 'thin'. A port is an orifice connecting two chambers as shown in Figure 2. The transfer of air mass is derived using thermodynamic properties of air considering the compression and the expansion stages of actuation as isentropic, hence the heat transfer coefficient for both the processes are assumed to be equal to specific heat ratio of air $n = 1.4$ [5].

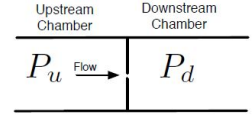

**Fig. 2:** The thin port model assumption

The pressure dynamics inside a chamber of a pneumatic cylinder can be described by the equations 1 - 5, these equations model the rate of change of pressure in the chamber as a function of the air mass flow $\dot{m}$, inlet and exhaust port area $a_c$ and $a_a$ respectively, volume of the chamber $v$ and the time derivative of the volume $\dot{v}$. The mass flow from the compressor is controlled using a proportional valve and is described by the equation 1. In this equation, the air mass flow $\dot{m}$ depends on the port area and $f(p_u, p_d)$ where $p_u$ is the upstream pressure and $p_d$ is the downstream pressure. Equation 2 and 3 describes the $f(p_u, p_d)$. The behaviour of the pressure dynamics is dependent on the ratio of $P_u/P_d$. Above a certain value, the pressure dynamics becomes highly non linear and the air flow is referred to as choked flow. Equation describes the total mass flow into the system. $P_c$ is the compressor pressure, $P_r$ is the atmospheric pressure and $p$ is current pressure inside the chamber.

The terms $n$, $Rs$, $\alpha$, $\beta$, $k$, $\theta$ are well defined physical constants and the values are mentioned in the appendix.

$$\dot{m} = a f(p_u, p_d) \qquad (1)$$

$$f(p_u, p_d) = \begin{cases} z(p_u, p_d) & p_u \geq p_d \\ -z(p_u, p_d) & p_u < p_d \end{cases} \qquad (2)$$

$$z(p_u, p_d) = \begin{cases} \alpha p_u \sqrt{\left(\frac{p_d}{p_u}\right)^{\frac{2}{k}} - \left(\frac{p_d}{p_u}\right)^{\frac{k+1}{k}}} & p_u/p_d \leq \theta \\ \beta p_u & p_u/p_d > \theta \end{cases} \qquad (3)$$

$$\dot{m} = a_c f_p(P_c, p) - a_a f_p(p, P_r) \qquad (4)$$

$$\dot{p} = \frac{n}{v}(R_s T \dot{m} - p \dot{v}) \qquad (5)$$

### B. Thin port Model With Leak

The thin-port model presented above does not take leakage from the cylinders into account. To account for this, we consider that the leak is through an equivalent orifice area from the cylinder. We use the same thin port model to predict the mass flow from the chamber. This additional air flow out of the cylinder chamber can then be incorporated into the model. In these equations 6 - 8. It is important to note that $p_u$ is actually the chamber pressure and $p_d$ is atmospheric pressure. $a_l$ represents the leakage area. The rest of the constants are the defined in the appendix and they are the same values as the ones used in the thin port pressure dynamics model.

$$m_{Leak} = a_l f(p_u, p_d) \quad (6)$$

$$f(p_u, p_d) = -z(p_u, p_d) \text{ when } p_u < p_d \quad (7)$$

$$z(p_u, p_d) = \begin{cases} \alpha p_u \sqrt{\left(\frac{p_d}{p_u}\right)^{\frac{2}{k}} - \left(\frac{p_d}{p_u}\right)^{\left(\frac{k+1}{k}\right)}} & p_u/p_d \leq \theta \\ \beta p_u & p_u/p_d > \theta \end{cases} \quad (8)$$

The effective mass flow in the cylinder described in equation 4 can be modified to equation 9.

$$\dot{m} = a_c f_p(P_c, p) - a_a f_p(p, p_r) - a_l f(p, P_r) \quad (9)$$

IV. FACTORS THAT AFFECT PRESSURE DYNAMICS

Pressure dynamics of air is highly non linear and has a large time delay due to the slow propagation of air pressure waves through the system. So, it is important to study the factors that might aggravate these effects. For example, valve deadzone adds more non-linearity to the system, length of the connecting tubes introduces proportional amount of delay into the system and leakage in the cylinders directly affects the air mass flow into the cylinder chamber. In this section we discuss these important factors.

*A. Inlet and Exhaust port areas*

These are controlled by the valve and have a direct effect on the mass flow into and out of the cylinder chamber. The rate of change in pressure inside the cylinder depends on these areas. The change in the area values with respect to the control is non-linear. This non-linearity has to be understood to predict the air flow from the valves.

*B. Valve leakage and deadzone*

Leakage in valves is the undesirable air flow through the inlet and exhaust ports. This has a significant effect in the pressure dynamics inside the cylinder. In an ideal valve, the air flow through the valve at zero command signal should be zero, but the flow rate we measured using a flow meter was 1 $l/min$. The valves also have a control range around midpoint in which no change in the air flow takes place, this is called the valve deadzone. For a small range in the control signal, the valve essentially does not control the air flow. This gives rise to a non linear behaviour.

*C. Volume of cylinder*

The over all volume of the cylinder has an important influence on the pressure response inside the chamber (Equation 5). Note that pressure dynamics is inversely proportional to the effective volume. On one hand where the pressure dynamics is known to be notoriously slow, the pressure dynamics of a fully retracted cylinder (approx. volume $0.1 cm^3$), exhibits pressure change on the time scale of 10 microseconds. A cylinder with smaller volume and volume fluctuations have a much faster pressure response than the one with large volumes and volume fluctuations.

*D. Leakage from cylinders*

Air leakage from the pneumatic cylinders also has a significant effect on pressure dynamics inside the cylinder. Some types of pneumatic cylinders tend to leak more than others, this depends on the type of seal that is used between the cylinder piston and bore. Cylinders that have a rubber seal have lesser leakage, however these cylinders have larger friction. Anti-stiction/friction cylinders overcome the effect of friction by increasing the gap between the piston and the cylinder bore (air-seal), this results in higher leakage.

*E. Delay from connecting tubes*

Delay in the pressure dynamics is affected by the length of the connecting tubes. The longer the tube, more time it takes for the air pressure waves to travel the entire distance of the tube and hence a delay is introduced to the system. The connecting tube also introduce pressure loss because of the friction factor that increases tube resistance.

*F. Input Pressure fluctuations*

The source of the compressed air is not always at constant pressure. While actuating the adroit hand, around 40 cylinders are drawing compressed air, hence pressure fluctuations from the source cannot be avoided. The pressure variations in the source has to be accounted for to increase the accuracy of the model.

V. HARDWARE OVERVIEW AND CONTROL

AIRPEL anti-stiction cylinders are used in the actuation system of Adroit. Based on the joint range and torque requirements, different cylinders varying in stroke length and bore area are used. The Bio-mimetic hand is actuated using AIRPEL M9D37.5 (AIR37) which has a bore diameter of $9mm$ and stroke length of $37.5mm$ because of the small range of motion required for the finger joints. Whereas Adroit robot arm is actuated using AIRPEL cylinders of diameter $24mm$, for high torque requirements, and of varying stroke as required at each joint. All the cylinders are linear pneumatic actuators. We test our model on two AIRPEL cylinders – AIR37 and AIR200 (Part number: M9D200. Diameter: $9mm$. Stroke length: $200mm$). Both cylinders have same diameter and leak properties, but different stroke length and hence different volumes (Table I). We chose these cylinders to compare the effect of the size of the cylinder on the model performance.

2-dof robot is built using different pneumatic actuators, one is a linear double actuation pneumatic cylinder SMC CQ2A32-25DC (SMC), and the other is a rotary double actuation pneumatic cylinder, PRNA20S-180-45 (PRN). These cylinders were chosen to test our model's performance on cylinders that have different leakage rates than the AIRPELs. The cylinder characteristics are mentioned in Table I.

FESTO MPYE-5-1/8-LF-010-B valves are used to control all the pneumatic actuators. Valves with different flow characteristics and deadzone reagions, shown in Figure 4, are chosen to study the effect of these on pressure dynamics. We used connecting tubes that are $2mm$ in diameter. The maximum compressor pressure used was 60 $kPa$, and to address the issue of fluctuating source pressure we used a pressure sensor to monitor the actual pressure from the source.

| Cylinder | Part number | Volume($m^3$) | Characteristics |
|---|---|---|---|
| AIR37 | AIRPEL M9D37.5 | 2.3856e-06 | High leak, small volume |
| AIR200 | AIRPEL M9D200 | 1.2723e-05 | High leak, large Volume |
| SMC | SMC CQ2A32-25DC | 2.0106e-05 | Low leak, large Volume |
| PRN | PRNA20S-180-45 | 3.50e-06 | Low leak, small volume |

**TABLE I:** Cylinder Characteristics

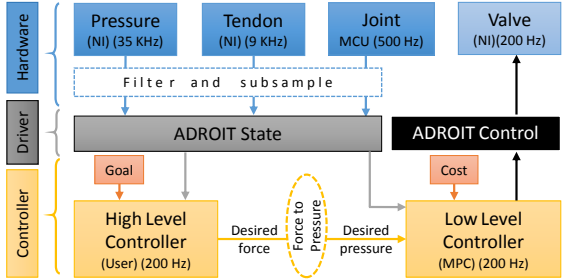

**Fig. 3:** The model based controller architecture being developed for Adroit Manipulation Platform

The pressure inside the cylinder unit is observed using a (SMC PSE540-IM5H3) pressure sensor. The cylinder piston stroke length using a magnetic length sensor (SICK MPS-032TSTU04). In case of the 2-dof robot, the piston stroke position is inferred from the joint angle sensor readings because the cylinders are non-magnetic in nature. The pressure sensors are sampled at 32KHz and the length sensors are sampled at 9Khz. High frequency components of the sensor readings are filtered out, using low pass filters, before they are made available for use. High sampling rate allows us to perform data filtering without introducing significant delays.

*A. Control Architecture*

The overall control architecture is illustrated in the Figure 3. The three main components of the architecture are $(a)$ Hardware: which consists of all physical components like sensors, control valves and electronics, $(b)$ Driver : where the data from the sensors are calibrated and assembled as states and controls, $(c)$ Controller : which uses the states as an input and sends the desired controls to the valves. The controller itself operates at two levels. At the beginning of each control cycle (running at 200Hz) using the current state the high level controller submits a desired torque profile to the low level controller pretending that robot is driven using a ideal torque actuator. The "force to pressure" module accounts for the transmission details (tendon routing, moment arms, dry friction etc.) to map desired joint force/torques to desired actuator pressures. The low level controller subsumes the complexities of the pneumatics and abstracts out a simple torque actuator to the high level controller for planning purposes. The low level controller (detailed in [3]) leverages online trajectory optimization techniques and pneumatic models to handle the complications of the pneumatic actuation in order to execute the request submitted by the high level controller.

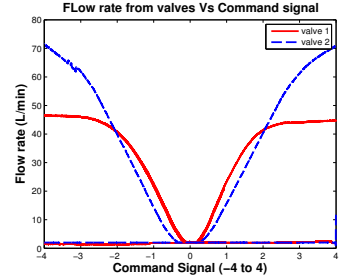

**Fig. 4:** This Figure shows flow measurement from two different valves. Because of the deadzone, the Valves have no response for a small range around the zero input command. The response of the two valves differ significantly.

## VI. EXPERIMENTATION AND RESULTS

In this section, we describe the experimental method used to identify the parameters of thin port model. First, to understand the flow characteristics we measure the air flow from the valves using a flow meter. Second, we measure the pressure response from pneumatic cylinder for a particular set of command signals. We use this data to optimize for the parameters. The experimental method is explained in detail in the next few paragraphs.

*A. Valve Characterization*

As described in the previous section, pressure changes inside a cylinder chamber is a function of the air mass flow. Characterizing this flow from the valve is crucial in understanding the pressure changes inside the chamber. Equation 1 describes the mass flow through an area $a$, the valves control this area by using solenoid actuation that moves a spool when a control input is applied. The control range of the valve is $0 - 10V$, we map this to a $-5\ to\ 5$ range for easy representation. A command signal $-5\ to\ 0$ opens the exhaust port and a command signal $0\ to\ 5$ opens the inlet port of the valve. To understand this behaviour of the valve as a function of the control input, we used a standard air flow meter (SFAB-200U-HQ8-25V-M12 from FESTO). The flow measured here is in $l/min$. The Figure 4 illustrates the data obtained from the flow sensor. It is important to note that the valves are far from ideal and there is some flow in both the directions as the command input is changed. The flat region around the zero command in Figure 4 is the deadzone in the valve.

The bandwidth of the valve is dependent on the amplitude of the spool displacement, and for a full length displacement of the spool, the bandwidth is 125Hz. Since the bandwidth of the pneumatic system is much lower than 100Hz, the dynamics of the valve itself can be neglected while modelling the pressure dynamics.

*B. Optimization*

Three parameters need to be identified in order to implement the thin port model for pressure predictions, first the mapping from command signal to the inlet and exhaust port area $a_c$ and $a_a$ respectively, the nature of this function is illustrated in Figure 4 where we can see that the there is always constant small flow in both the directions due to leakage. The second parameter that needs to be identified is the volume of the cylinder $v$. In addition to this, we also need to estimate the effective leak area $a_l$ described earlier. These parameters can be identified using data driven optimization techniques. We collect the pressure response data by first fixing the cylinder at a particular volume. Then we apply random step commands to the valve. By minimizing the error between the measured rate of change in pressure and the rate of change in pressure predicted by model we estimate values for the variables $a_c, a_a, a_l, v_o$. This is done for a whole range of control inputs. By doing this, we obtain a mapping of the area with respect to the control signal. If this experiment is repeated at different fixed volumes, then we can also build a function that maps sensor readings to volume of the chamber inside.

| valve | offset | a | b | c | d |
|---|---|---|---|---|---|
| Valve 1 inlet area | 2.9274e-08 | 5.501e-07 | 3.539 | 1.564 | 0.12 |
| Valve 1 exhaust area | 2.6076e-08 | 6.079e-07 | 3.149 | -1.365 | -0.1 |
| Valve 2 inlet area | 2.627e-08 | 8.731e-07 | 4.029 | 0.929 | -0.1505 |
| Valve 2 exhaust area | 2.964e-08 | 8.659e-07 | 3.942 | -0.9816 | 0.1584 |

**TABLE II:** Optimized values for the mathematical model of inlet and exhaust ports of the valves. Illustrates how each valve has different characteristics

This optimization problem can be set up to identify the desired area and volume parameters. In equation 10, $\dot{P}_m$ is the measured rate of change of pressure inside the cylinder. The other terms are described in section III.

$$a_c, a_r, v, a_l = \underset{a_c, a_r, v, a_l}{argmin} \{\dot{P}_m - \frac{nR_sT}{v}(a_c f(P_c, p) - (a_a + a_l)f(p, P_r))\} \quad (10)$$

The built-in optimization tool box in MATLAB was used to solve the optimization problem described above.

*1) Inlet and Exhaust areas:* The identified input and exhaust valve areas are shown in Figure 5(a). It is important that the function that represents the area as a function of the command signal be a continuous function because any discontinuity will introduce undesired non-linear behaviour. We chose the gompertz function, given in equation 11, to represent the area function and used standard matlab curve fitting tool to obtain the parameters $offset, a, b, c\ and\ d$. The values obtained for different valves are presented in table II. $cmd$ is the control input to the valve.

$$area = offset + (a)exp^{(-b)exp^{(-c)(cmd+d)}} \quad (11)$$

*2) Volume:* The volume of the cylinder during actuation is extremely important in building an accurate model for the pressure dynamics. The volume inside the pneumatic cylinder can be estimated by sensing the position of the piston and multiplying it with the bore area, but this does not include the dead volume in the chamber. By optimizing for the volume, we are essentially estimating the dead volume of the cylinder also. Based on the optimized values, we represent the volume as a function of the sensor readings directly. As expected, the volume function obtained through optimization was a linear function. This is illustrated in Fig. 5(b).

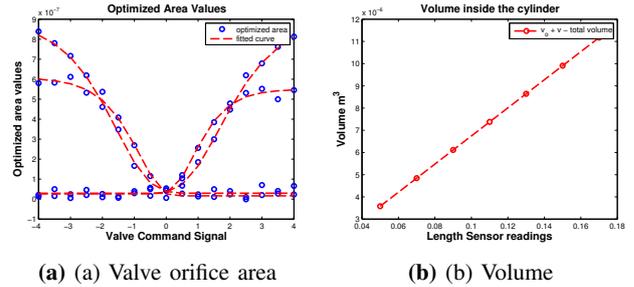

**(a)** (a) Valve orifice area  **(b)** (b) Volume

**Fig. 5:** The Figure (a) shows the optimized values for two different types of valves. The difference in the flow characteristics is captured by optimization, hence taking into account the valve leakage. Figure (b) shows the optimized volume function

| Cylinder | RMSE | Percent of pressure range |
|---|---|---|
| SMC | 1.476e+04 | 3.2 |
| PRN | 1.369e+04 | 3.042 |
| AR200 | 5.161e+04 | 11.4 |
| AR37 | 1.370e+04 | 3.04 |

**TABLE III:** The RMSE values for the pressure predictions are given in this table.

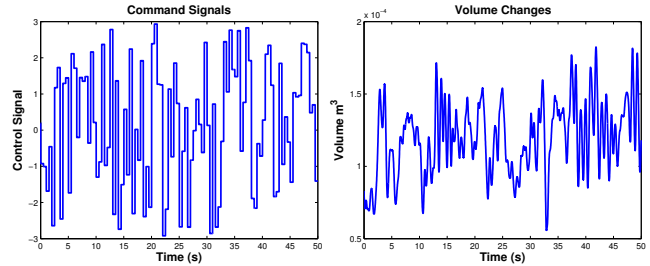

**Fig. 6:** (Left): An example of the type of random control signal used for prediction. (Right): Volume change the cylinder is subjected to during data collection.

*3) Leakage area:* For Airpel cylinders, the optimized value of the orifice diameter through which leakage is assumed to take place was found to be $1.9665e-04m$.

## VII. RESULTS

*1) Pressure Predictions:* In this section, the pressure modelling results are investigated on actual hardware. To illustrate the effectiveness of the identified model, we apply fast changing random commands to the valves and manually provide aggressive volume changes to the cylinders. Then we compare the pressure changes predicted by the identified model to the measured pressure changes during the hardware experiment. Pressure modelling results on all four cylinders are presented. An example of the control signal and volume change during the data collection is illustrated in Figure 6. It is important to note that the pressure is measured in Pascals (Pa).

Figures 7 and 8 illustrates the pressure modelling achieved from PRN and SMC cylinders.

Figures 9 and 10 show the pressure modelling achieved in AR37 and AR200 cylinders. In this Figure, the light blue line represents the volume change in the cylinder. As the figures illustrate the identified model is robust and is able to match the pressure changes measured inside the cylinder. The RMSE values obtained are shown in Table III

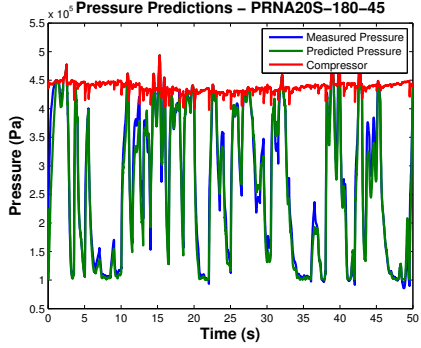

**Fig. 7:** Pressure predictions for PRN cylinder

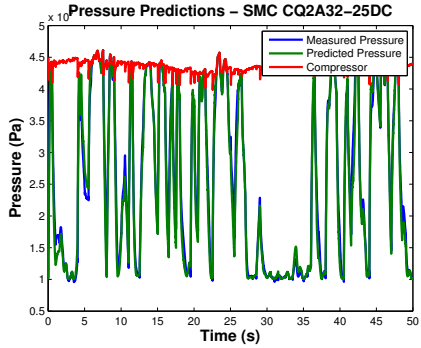

**Fig. 8:** Pressure predictions for SMC cylinder

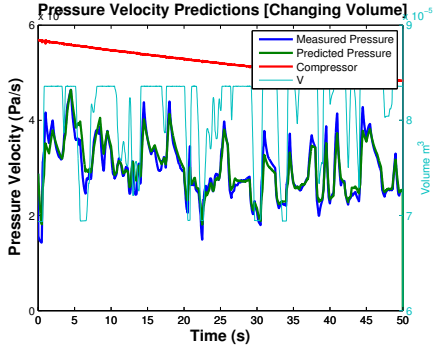

**Fig. 9:** Pressure predictions for AR37 cylinder

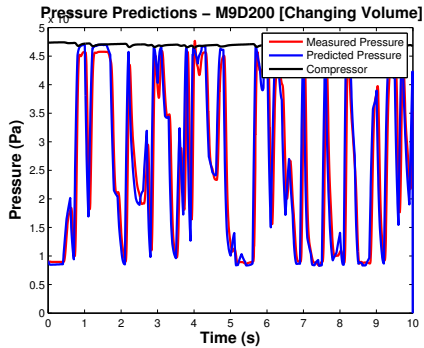

**Fig. 10:** Pressure predictions for AR200 cylinder

## VIII. DISCUSSION AND CONCLUSION

In this section, we analyse the performance of the model based on the error in pressure predictions on the different cylinders. We discuss how our model is affected by parameters such as the leakage and deadzone in the valves, leakage from cylinders and large volume. We also highlight some of the shortcomings of the model and the reasons for them.

The model based low level controller leverages the pneumatic model to look through the planning horizon and selects the optimal policy that results in effective pressure predictions. So, we are most interested in the prediction of the model for a short duration of about $2s$ into the future. We initialize the model to initial measured pressure and let the model predict the pressure changes for the next $2s$. To check the model performance, we take the mean of prediction error for multiple two second trajectories and assess them based on comparing the error as a percentage of the working pressure range.

From the results we obtained, the pressure predictions were well under $4\%$ of the working pressure range for all the cylinders except AR200. The reason for the relatively poor performance of the model with AR200 is a combination of the leakage and larger volume.

### A. Effect of Volume

Larger volume has an adverse effect on the performance of the model because the $v$ term in the thin port model has more impact on the overall pressure dynamics of the system. This effect is illustrated in the Figure 11(a), Both Airpel cylinders suffer from leakage, but it is much harder to predict the pressure changes in the AR200 with a larger volume. Figure 11(a) also shows the comparison between SMC and PRN actuators. The prediction error in SMC is more because to its larger volume.

### B. Effect of valve deadzone

The valve characteristics is hard to model around the zero command signal because of the non linear behaviour introduced by the valve deadzone around that control region. This leads to some error in prediction for control signals in this region. This is shown in the Figure 12(b), the error obtained by subjecting the pneumatic system to a random fast changing control signal between the range -0.1 to 0.1 is larger than the error obtained for a random control input between the range -3 to 3. This is an important factor when the model is implemented with a controller, because when the pressure in the cylinder has to be maintained at a particular value, the controller operates in this small range around the deadzone. However, the errors are still less than $4\%$ of the working pressure range, so the area function comes close to modelling this region.

### C. Effect of leakage in the cylinders

Figure 11(a) compares the performance of the identified model for SMC, PRN, AR37 and AR200 cylinders, as mentioned earlier, the leakage from SMC and PRN cylinders is negligible compared to the leakage from AR37 and AR200 cylinders. Hence, it is not surprising that the errors in SMC and PRN are lesser. However, Modelling the leak based on our assumption that the leak takes place through an equivalent

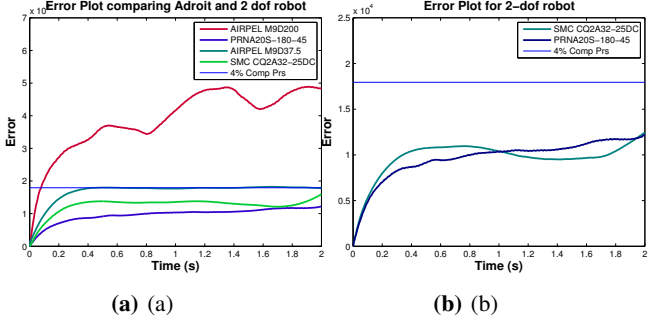

**Fig. 11:** Fig (a) compares the error in prediction from all the four cylinders. Airpel cylinders that suffer from more leakage have larger error. Fig (b) compares the errors in SMC ans PRN cylinders.

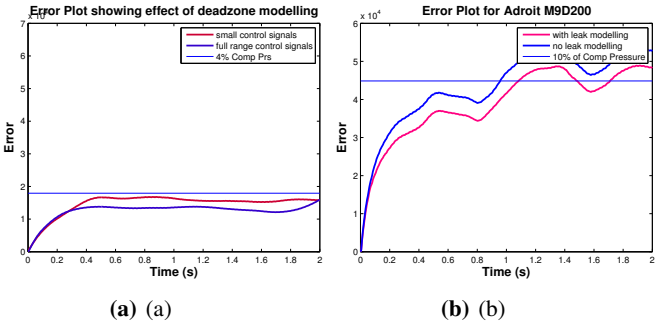

**Fig. 12:** Fig (a) illustrates the reduction in error by modelling the leak using the proposed method. Figure (b) illustrates the difficulty in modelling the valve around the zero control region because of non linear behaviour due to valve deadzone. The red line is the error in prediction when the control range is close to zero but the blue line is the prediction error over a larger control range. It is easier to estimate the inlet and exhaust port areas for larger control signals.

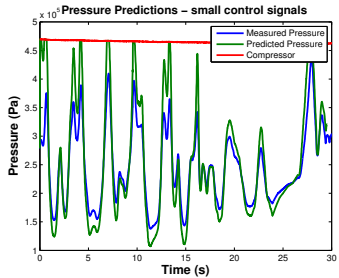

**Fig. 13:** This Figure shows the increased error in pressure predictions by the identified model for control signals close to zero.

orifice area from the cylinder does reduce the error. This is illustrated in 12(a).

In conclusion, a method for identifying a pneumatic model for Adroit manipulation platform was presented. The effects of valve leakage, valve deadzone, leakage in cylinders and input pressure variations are taken into account to increase the accuracy of the model. Pressure predictions achieved for the different cylinders satisfy the accuracy and robustness requirement for implementing it in a controller. We also highlighted the factors that determine the difficulty in modelling a pneumatic system by comparing the errors in the prediction.

## IX. FUTURE WORK

In future work, the effect of connecting tube length will be included in the model to improve the accuracy. The identified model presented in the paper has already been used in our other work to track pressure inside the cylinder by implementing the control architecture mentioned in this work [3]. Further work will include improving the performance of the controller.

## APPENDIX

The physical constants are given by:

$$\alpha = C \sqrt{\frac{2M}{Z\,R\,T}\frac{\kappa}{\kappa-1}} \qquad \theta = \left(\frac{\kappa+1}{2}\right)^{\frac{\kappa}{\kappa-1}}$$

$$\beta = C \sqrt{\frac{\kappa M}{Z\,R\,T}\left(\frac{2}{\kappa+1}\right)^{\frac{\kappa+1}{\kappa-1}}}$$

| Gas Molecular Mass | $M$ | 0.029 for air, $Kg/mol$ |
| Temperature | $T$ | $K^\circ$ |
| Universal Gas Constant | $R$ | 8.31 $(Pa \cdot m^3)/(mol\,K^\circ)$ |
| Discharge coefficient | $C$ | 0.72, dimensionless |
| Compressibility Factor | $Z$ | 0.99 for air, dimensionless |
| Specific Heat Ratio | $\kappa$ | 1.4 for air, dimensionless |
| Mass Flow | $\dot{m}$ | $Kg/s$ |
| Pressure | $p$ | $Pascals$ |
| Area | $a$ | $m^2$ |

**TABLE IV:** Parameters and units of the thin-plate port model.